\documentclass{article}
\usepackage{spconf,amsmath,amssymb,graphicx,hyperref}
\usepackage{booktabs}        
\usepackage{xcolor}          
\usepackage{pifont}          
\usepackage{tikz}            
\usetikzlibrary{calc}        
\usetikzlibrary{arrows.meta} 

\title{DuplexDrama: A Synthesized Dialogue Dataset with Scenarios, Full-Duplex Behaviors, Expressive Speech, and Sound Events}

\makeatletter
\def\@maketitle{\newpage
 \null
 \vskip 2em \begin{center}
 {\large \bf \@title \par} \vskip 1.5em {\large \lineskip .5em
\parbox{\textwidth}{\centering\@name\par\@address}%
 \par} \end{center}
 \par
 \vskip 1.5em}
\makeatother
\name{Qingxiang Guo\sthanks{Corresponding author: \texttt{guoqingxiang@zuoyebang.com}},
      Wenke Fan,
      Shuofeng Zhao,
      Dawei Yang,
      Zhiyang Zhou,
      Yingxin Shang, \\
      Hongwei Cai,
      Zhou Wang,
      Weixu Wang,
      Lin Yang,
      Shuran Zhou,
      and Yang Song}
\address{Zuoyebang Education Technology}

\begin{document}
\sloppy                    
\maketitle

\begin{abstract}
We present DuplexDrama, the first synthesized spoken dialogue dataset that simultaneously covers four dimensions: (i) complete persona and scenario settings; (ii) three full-duplex behaviors (interruption, backchannel, incomplete); (iii) expressive speech with persona-aligned emotion labels; and (iv) script-aware sound events. DuplexDrama is built via a 4-stage pipeline; quality validation on both scripts and synthesized audio confirms its quality. We have produced more than 2{,}000 hours audio data with a 64-voice timbre pool spanning 13 personas and 5 age buckets; 3.8\% of all turns carry at least one full-duplex behavior. This data has been validated through internal full-duplex model training~\cite{adaptduplex}. We will release a curated subset of 6{,}400 bilingual dialogues (800~h, Chinese $\sim$500~h + English $\sim$300~h) to advance full-duplex spoken dialogue model research. Data samples are available at our demo page\footnote{\url{https://dunjie5465.github.io/duplexdrama-demo/}} and LLM-judge evaluation prompts will be released with the dataset.
\end{abstract}


\section{Introduction}
\label{sec:intro}

\begin{figure}[t]
\centering
\small
\input{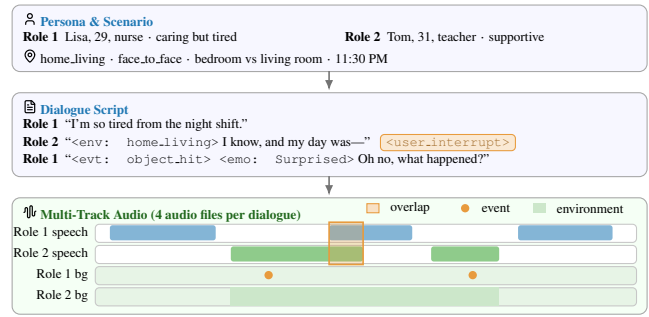}
\caption{One DuplexDrama sample: persona and scenario metadata, a tagged
dialogue script, and four aligned audio tracks in which tagged interruptions
surface as real acoustic overlap. Swapping Role~1 and Role~2 yields two
training samples per dialogue, doubling the usable data.}
\label{fig:sample}
\end{figure}

The emergence of large language models has enabled conversational AI systems
that listen and speak simultaneously~\cite{lu2026full}---ranging from
open-source full-duplex models~\cite{moshi} to commercial
systems~\cite{gpt4o}.
Yet training such systems requires full-duplex dialogue data that is
severely scarce.
Existing TTS-synthesized dialogue corpora~\cite{sdf} are
predominantly clean read speech or simple turn-taking exchanges.
Full-duplex benchmarks~\cite{fdb} target evaluation only and remain
small in scale.
Natural corpora spanning thousands of hours~\cite{fisher,candor,duplexconv,openyap1k} ship authentic overlap and backchannel in the audio, yet ship without scripted scenarios, rich persona attributes, or full-duplex behavior annotations.
Recent persona-conditioned full-duplex models~\cite{personaplex} target
voice and role control, but do not release public training corpora.
Moreover, background noise and sound events can provide multimodal
information beyond speech, but they are not used in synthesizing dialogue
data yet.
To our knowledge, no public dataset jointly covers (i)~complete persona
and scenario annotations---which condition the model's response style;
(ii)~systematic coverage of the three full-duplex behaviors
(interruption, backchannel, incomplete); (iii)~expressive speech
with persona-aligned emotion labels; and (iv)~script-aware sound event
injection. PersonaPlex~\cite{personaplex} demonstrates persona-conditioned
full-duplex conversational modeling; complementing this line of
work, DuplexDrama provides a fully open dataset covering all four
dimensions.
We fill this gap with DuplexDrama, a synthesized spoken dialogue corpus.
Fig.~\ref{fig:sample} shows one complete sample.

DuplexDrama is built via a four-stage pipeline:
(§2.1) Persona and scenario generation with consistency constraints;
(§2.2) Script generation with full-duplex behaviors, emotion tags and sound event tags;
(§2.3) IndexTTS2-based~\cite{indexTTS2} expressive speech synthesis with a
64-voice timbre pool spanning 13 personas and 5 age buckets, followed by audio assembly;
(§2.4) Script-aware event injection at the trigger word via word-level timestamps, with controlled SNR.
A quality validator (§2.5) then discards low-quality dialogues, retaining only the high-quality subset for release.

We make the following contributions:
\begin{itemize}
\item A four-stage pipeline that constructs a synthesized spoken
dialogue dataset jointly covering persona and scenario, three
full-duplex behaviors, persona-aligned expressive speech, and
script-aware sound events, followed by a validator to assert quality.
\item An evaluation protocol leveraging dual LLM judges (DeepSeek-v4.1-pro and Gemini-3.1-pro-preview) for cross-checking, with evaluation prompts to be released alongside the dataset, to enable reproducible script-rationality assessment (Section~\ref{sec:experiments}).
\item A full open release of 6{,}400 bilingual dialogues (800~h),
together with the audio asset bank (600 environmental audios and
2{,}203 sound-event audios) and LLM-judge evaluation prompts, to
support reproducible full-duplex spoken dialogue model research.
\end{itemize}

\section{Pipeline}
\label{sec:pipeline}

\begin{figure*}[t!]
\centering
\small
\input{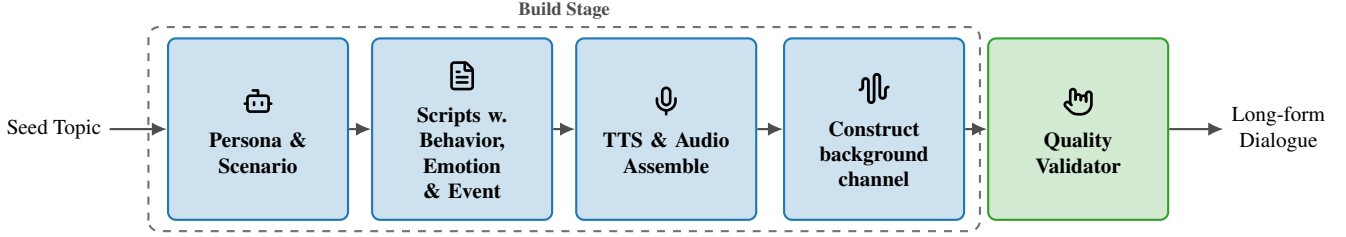}
\caption{DuplexDrama pipeline generating persona, behavior/emotion/event-tagged scripts, dual-track long-form speech waveforms with full-duplex behaviors surfaced as acoustic overlap, and script-aware event mixing, followed by a quality validator.}
\label{fig:pipeline}
\end{figure*}

Figure~\ref{fig:pipeline} overviews our four-stage pipeline. Stage 1 generates persona and scenario tuples under consistency constraints. Stage 2 takes these validated tuples and generates dialogue scripts annotated with three tag families---full-duplex behavior, emotion, and sound-event tags. Stage 3 synthesizes audio via IndexTTS2~\cite{indexTTS2} with voice--emotion conditioning and concatenates per-utterance segments in script order. Stage 4 constructs a background channel for each speech channel with both environmental noise and sound events. The pipeline produces long-form dialogue audios. A quality validator then filters each dialogue; failures are discarded before release.

\subsection{Persona and Scenario Generation}
\label{ssec:persona}

We first construct two complementary seed pools via
DeepSeek-v4.1-pro~\cite{deepseekai2026deepseekv4}: a persona pool covering
24 macro categories and 134 sub-types organized by
social function, and a topic pool covering
46 macro categories and 404 sub-types organized by
narrative scenario. From each pool we draw a seed, then DeepSeek-v4.1-pro
expands the seed into a dialogue setup that couples two persona tuples\footnote{name, gender, age, occupation, personality, role.}
with a scenario specification\footnote{scene, tone, narrative\_type, scene\_medium, an
event chain of three progressing events, and a segmented emotional arc.}
under consistency constraints.

\subsection{Dialogue Script Generation}
\label{ssec:script}

In Stage~2, the LLM follows the validated persona--scenario tuple and
generates the dialogue script annotated with three orthogonal tag
families: (i) full-duplex behavior tags (\texttt{<user\_interrupt>},
\texttt{<user\_backchannel>}, \texttt{<user\_incomplete>}); (ii) proper
emotion tags (\texttt{[Neutral]}, \texttt{[Happy]}, \texttt{[Angry]},
\texttt{[Sad]}, \texttt{[Whispering]}, \texttt{[Hesitant]},
\texttt{[Surprised]}) aligned with IndexTTS2; (iii) environmental noise
and sound event tags referencing the audio bank.
In addition, the LLM is prompted to produce spoken-style dialogue.
Gemini-3.1-pro-preview~\cite{gemini31pro} is used in this stage.

\subsection{Full-duplex Dialogue Synthesis}
\label{ssec:voice}

We synthesize expressive speech via IndexTTS2~\cite{indexTTS2} since it
supports disentangled control of speaker timbre and emotion.
Our voice pool contains 64 speakers spanning 13 core personas
and 5 age buckets. Most prompt audios are generated by MOSS-Audio~\cite{mossAudio}, while a small fraction
curated from natural speech snippets.
The seven-class emotion labels are defined at Stage~2 (§2.2),
where the LLM is constrained to draw only from these classes in scripts.
A speaker prompt audio and an emotion tag jointly condition IndexTTS2 at synthesis time.

After obtaining per-utterance audio, we first run forced alignment (FA)~\cite{qwen3ASR}
on all utterances to obtain word-level timestamps.
We then assemble the utterances by turn order into two parallel tracks for both speakers.
Each track only carries its own speaker; during the other speaker's
turns, silence is inserted to preserve speaker-switch boundaries.
This yields two long-form speech waveforms.

Full-duplex behaviors are jointly realized with the former steps.
Specifically, \texttt{<user\_incomplete>} tags are replaced by literal ``\ldots'' inside
sentence, which IndexTTS2 renders as in-sentence pauses during synthesis.
During audio assembly, for an \texttt{<user\_interrupt>} tag, the corresponding utterance
fades out briefly while the other turn is brought in, producing acoustic overlap;
for an \texttt{<user\_backchannel>} tag, the corresponding utterance is time-aligned and overlaid
onto the other track which continues speaking, yielding simultaneous speech.

\subsection{Background Channel Realization}
\label{ssec:events}

Stage~2 (§2.2) already places environmental noise and sound-event tags at semantically meaningful
positions in the script. Stage~4 realizes those tags into audio.

\begin{figure}[htb]
\centering
\small
\input{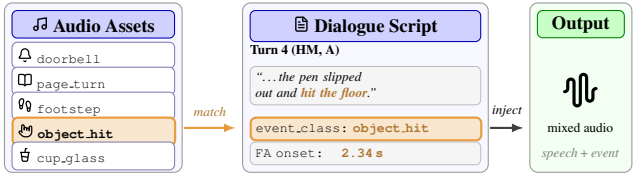}
\caption{Sound-event realization: assets from the audio bank are matched against the \texttt{event\_class} tag emitted by Stage~2 (§2.2), aligned to the trigger word's onset by forced alignment of the synthesized speech, and mixed in at the configured SNR.}
\label{fig:event-injection}
\end{figure}

We construct two background channels per dialogue, each paired with one
speech channel and combining environmental noise and sound events.
Environmental noise is drawn from a self-built bank of 7 scene categories
and 600 long-duration clips. Each noise instance persists for several
turns, which is decided at script generation time.
Sound events are drawn from a separate bank of 52 categories and
2{,}203 short-duration clips. Each event tag is matched to an
asset and inserted at the FA-aligned trigger position (Fig.~3).
Both background noise and sound-event loudness are controlled by
configurable SNR. Finally, each speech channel pairs with a time-aligned
background channel. In general, training a full-duplex model only requires the background channel at the input (HM) side.

\textit{All audio assets are drawn from publicly accessible sources and
processed strictly for non-commercial, academic research purposes.}

\subsection{Quality Validation}
\label{ssec:validation}

We employ a dual-LLM (DeepSeek-v4.1-pro~\cite{deepseekai2026deepseekv4} and
Gemini-3.1-pro-preview~\cite{gemini31pro}) framework to evaluate scripts on
three dimensions: asset rationality of the manually curated audio bank
(judged by the single Gemini-3.1-pro-preview multimodal model),
tag rationality in scripts, and script-scenario consistency, with the
latter two cross-checked by both LLMs.
The synthesized audio is then evaluated along four objective metrics: WER
(Qwen3-ASR~\cite{qwen3ASR}), audio quality via UTMOSv2~\cite{baba2024utmosv2}
and NISQA~\cite{nisqa} (MOS), and SpkCons via SpeechBrain~\cite{speechbrain}
cosine similarity. Dialogues with SpkCons below 0.9 are discarded following~\cite{sdf}.

\section{Dataset Analysis}
\label{sec:experiments}

In this section, we analyse the dataset through intrinsic metrics
organized into three sub-sections.
Section~\ref{ssec:stats} reports dataset statistics;
Section~\ref{ssec:quality} reports quality validation;
Section~\ref{ssec:comparison} compares with prior corpora.

\subsection{Dataset Statistics}
\label{ssec:stats}

\begin{figure}[htb]
\centering
\small
%
%
%

\definecolor{cbBlue}{HTML}{1F78B4}

\begin{tikzpicture}[
    every node/.style={font=\tiny},
    grid/.style={dashed, draw=black!18, thin},
    axis/.style={->, draw=black!55, thin},
    bar/.style={fill=cbBlue!55, draw=cbBlue, thin,
        rounded corners=1.5pt},
    xlab/.style={font=\tiny, anchor=north east, inner sep=1pt,
        rotate=45, align=right}
]

\foreach \p in {10,20,30,40}{
    \draw[grid] (0.55, \p*0.05) -- (7.85, \p*0.05);
    \node[font=\tiny, text=black!55, anchor=east]
        at (0.50, \p*0.05) {\p\%};
}

\draw[axis] (0.55, 0.00) -- (0.55, 2.10);
\draw[axis] (0.55, 0.00) -- (7.95, 0.00);

\filldraw[bar] (1.03, 0.00) rectangle (1.59, 32.55*0.05);
\node[font=\tiny, anchor=south] at (1.31, 32.55*0.05) {32.55\%};
\node[xlab] at (1.31, 0.00) {home\_living};

\filldraw[bar] (2.07, 0.00) rectangle (2.63, 23.68*0.05);
\node[font=\tiny, anchor=south] at (2.35, 23.68*0.05) {23.68\%};
\node[xlab] at (2.35, 0.00) {quiet\_indoor};

\filldraw[bar] (3.11, 0.00) rectangle (3.67, 19.52*0.05);
\node[font=\tiny, anchor=south] at (3.39, 19.52*0.05) {19.52\%};
\node[xlab] at (3.39, 0.00) {office};

\filldraw[bar] (4.15, 0.00) rectangle (4.71, 11.40*0.05);
\node[font=\tiny, anchor=south] at (4.43, 11.40*0.05) {11.40\%};
\node[xlab] at (4.43, 0.00) {street\_outdoor};

\filldraw[bar] (5.19, 0.00) rectangle (5.75, 5.60*0.05);
\node[font=\tiny, anchor=south] at (5.47, 5.60*0.05) {5.60\%};
\node[xlab] at (5.47, 0.00) {public\_lobby};

\filldraw[bar] (6.23, 0.00) rectangle (6.79, 3.69*0.05);
\node[font=\tiny, anchor=south] at (6.51, 3.69*0.05) {3.69\%};
\node[xlab] at (6.51, 0.00) {cafe\_restaurant};

\filldraw[bar] (7.27, 0.00) rectangle (7.83, 3.56*0.05);
\node[font=\tiny, anchor=south] at (7.55, 3.56*0.05) {3.56\%};
\node[xlab] at (7.55, 0.00) {vehicle\_inside};

\end{tikzpicture}
\caption{Distribution of environment background tags across the seven
scene categories.}
\label{fig:env-dist}
\end{figure}
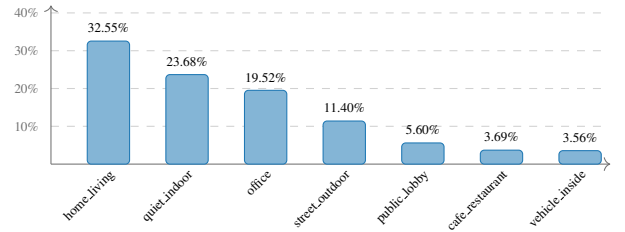

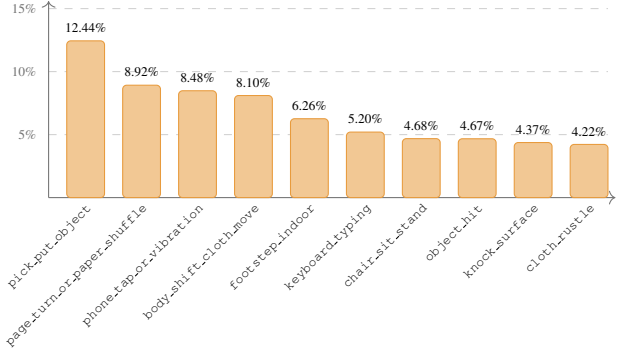
\begin{figure}[htb]
\centering
\small
%
%
%

\definecolor{cbOrange}{HTML}{E89B3D}

\begin{tikzpicture}[
    every node/.style={font=\tiny},
    grid/.style={dashed, draw=black!18, thin},
    axis/.style={->, draw=black!55, thin},
    bar/.style={fill=cbOrange!55, draw=cbOrange, thin,
        rounded corners=1.5pt},
    xlab/.style={font=\tiny\ttfamily, anchor=north east, inner sep=1pt,
        rotate=45, align=right}
]

\foreach \p in {5,10,15}{
    \draw[grid] (0.55, \p*0.1667) -- (7.95, \p*0.1667);
    \node[font=\tiny, text=black!55, anchor=east]
        at (0.50, \p*0.1667) {\p\%};
}

\draw[axis] (0.55, 0.00) -- (0.55, 2.60);
\draw[axis] (0.55, 0.00) -- (8.05, 0.00);

\filldraw[bar] (0.79, 0.00) rectangle (1.29, 12.44*0.1667);
\node[font=\tiny, anchor=south] at (1.04, 12.44*0.1667) {12.44\%};
\node[xlab] at (1.04, 0.00) {pick\_put\_object};

\filldraw[bar] (1.53, 0.00) rectangle (2.03, 8.92*0.1667);
\node[font=\tiny, anchor=south] at (1.78, 8.92*0.1667) {8.92\%};
\node[xlab] at (1.78, 0.00) {page\_turn\_or\_paper\_shuffle};

\filldraw[bar] (2.27, 0.00) rectangle (2.77, 8.48*0.1667);
\node[font=\tiny, anchor=south] at (2.52, 8.48*0.1667) {8.48\%};
\node[xlab] at (2.52, 0.00) {phone\_tap\_or\_vibration};

\filldraw[bar] (3.01, 0.00) rectangle (3.51, 8.10*0.1667);
\node[font=\tiny, anchor=south] at (3.26, 8.10*0.1667) {8.10\%};
\node[xlab] at (3.26, 0.00) {body\_shift\_cloth\_move};

\filldraw[bar] (3.75, 0.00) rectangle (4.25, 6.26*0.1667);
\node[font=\tiny, anchor=south] at (4.00, 6.26*0.1667) {6.26\%};
\node[xlab] at (4.00, 0.00) {footstep\_indoor};

\filldraw[bar] (4.49, 0.00) rectangle (4.99, 5.20*0.1667);
\node[font=\tiny, anchor=south] at (4.74, 5.20*0.1667) {5.20\%};
\node[xlab] at (4.74, 0.00) {keyboard\_typing};

\filldraw[bar] (5.23, 0.00) rectangle (5.73, 4.68*0.1667);
\node[font=\tiny, anchor=south] at (5.48, 4.68*0.1667) {4.68\%};
\node[xlab] at (5.48, 0.00) {chair\_sit\_stand};

\filldraw[bar] (5.97, 0.00) rectangle (6.47, 4.67*0.1667);
\node[font=\tiny, anchor=south] at (6.22, 4.67*0.1667) {4.67\%};
\node[xlab] at (6.22, 0.00) {object\_hit};

\filldraw[bar] (6.71, 0.00) rectangle (7.21, 4.37*0.1667);
\node[font=\tiny, anchor=south] at (6.96, 4.37*0.1667) {4.37\%};
\node[xlab] at (6.96, 0.00) {knock\_surface};

\filldraw[bar] (7.45, 0.00) rectangle (7.95, 4.22*0.1667);
\node[font=\tiny, anchor=south] at (7.70, 4.22*0.1667) {4.22\%};
\node[xlab] at (7.70, 0.00) {cloth\_rustle};

\end{tikzpicture}
\caption{Top-10 sound-event tags across five macro-classes, sorted descending by share.}
\label{fig:event-dist}
\end{figure}

We summarize the released corpus through four views: an overview
table (Table~\ref{tab:overview}), a full-duplex behavior-tag table
(Table~\ref{tab:fd-tags}), and two bar charts depicting the
distributions of environment background and sound-event
tags (Fig.~\ref{fig:env-dist}, Fig.~\ref{fig:event-dist}).

\begin{table}[htb]
\centering
\small
\caption{Dataset overview of the released corpus.}
\label{tab:overview}
\begin{tabular}{lr}
\toprule
\textbf{Metric} & \textbf{Value} \\
\midrule
Dialogues                  & 6{,}400 \\
Total audio (hours)        & 800 \\
Avg. dialogue length (s)   & 460 \\
Turns                      & $\sim$362k \\
Avg. turn length (s)       & 8 \\
Voice timbres              & 64 \\
\bottomrule
\end{tabular}
\end{table}

\begin{table}[htb]
\centering
\small
\caption{Full-duplex behavior tag counts and shares across all turns.}
\label{tab:fd-tags}
\begin{tabular}{lrr}
\toprule
\textbf{Behavior Tag} & \textbf{Count} & \textbf{Share} \\
\midrule
\texttt{<user\_interrupt>}   & 8{,}996 & 2.4\% \\
\texttt{<user\_backchannel>} & 2{,}455 & 0.6\% \\
\texttt{<user\_incomplete>}  & 2{,}271 & 0.7\% \\
\bottomrule
\end{tabular}
\end{table}

\subsection{Data Quality Validation}
\label{ssec:quality}

We evaluate each generated dialogue on the four objective audio
metrics defined in Section~\ref{ssec:validation}; script-tag
rationality is reported separately in Section~\ref{sssec:eventval}.
The results are listed as follows.

\subsubsection{Objective Audio Quality Metrics}
\label{sssec:objmetrics}

Note that both Word Error Rate (WER) and speaker consistency (SpkCons)
are performed on clean TTS audio (without background), since those two
metrics evaluate speech itself and should not be affected by additive
background noise. UTMOSv2 and NISQA are predicted on both the clean
TTS audio and the mixed (with background) audio as comparison.
Table~\ref{tab:objmetrics} reports the per-dialogue scores; the middle
column flags whether each metric is evaluated on clean, mixed, or both audio.
As expected, mixing the background costs UTMOSv2 only 0.05 but NISQA-MOS 0.47, since
UTMOSv2 targets the naturalness of synthetic speech while NISQA also
penalises additive noise; the speech itself is therefore left intact.
We attribute the lower-than-typical UTMOSv2 scores primarily to the
diverse persona stylings and emotion injection in our synthesized audio.

\begin{table}[htb]
\centering
\small
\caption{Per-dialogue \textbf{audio-side} objective metrics: WER / SpkCons / UTMOSv2 / NISQA. Middle column: \ding{55}~clean speech, $\checkmark$~mixed audio.}
\label{tab:objmetrics}
\begin{tabular}{lcc}
\toprule
\textbf{Metric} & \textbf{w. bg} & \textbf{Value} \\
\midrule
WER                  & \ding{55} & 1.8\% \\
SpkCons              & \ding{55} & 97.3\% \\
UTMOSv2              & \ding{55}/$\checkmark$ & 2.57 / 2.52 \\
NISQA --- MOS        & \ding{55}/$\checkmark$ & 3.65 / 3.18 \\
\bottomrule
\end{tabular}
\end{table}

\subsubsection{Script Rationality Validation via Dual LLM Judges}
\label{sssec:eventval}

We employ LLM judges to evaluate three dimensions: (i)~asset
rationality, (ii)~rationality of background tags, and (iii)~script-scenario
consistency. Dimensions (i) and (iii) are cross-checked by both
DeepSeek-v4.1-pro~\cite{deepseekai2026deepseekv4} and
Gemini-3.1-pro-preview~\cite{gemini31pro}; dimension (ii) is judged
by the single Gemini-3.1-pro-preview multimodal model since the
audio bank is manually curated.
Table~\ref{tab:eventval} reports average scores after filtering.

\begin{table}[htb]
\centering
\footnotesize
\caption{LLM validation results including asset rationality score of audio asset,
tag plausibility score and rationality score between dialogue settings and scripts.
All scores range from 0 to 5. (Env. for environment noise and Evt. for sound event.)}
\label{tab:eventval}
\resizebox{\columnwidth}{!}{%
\setlength{\tabcolsep}{3pt}%
\begin{tabular}{l ccccc}
\toprule
 & \multicolumn{2}{c}{\textbf{Asset rationality}} & \multicolumn{2}{c}{\textbf{Tag rationality}} & \textbf{Script} \\
\cmidrule(lr){2-3}\cmidrule(lr){4-5}
\textbf{Model} & \textbf{Env} & \textbf{Evt} & \textbf{Env} & \textbf{Evt} & \textbf{consist.} \\
\midrule
DeepSeek~\cite{deepseekai2026deepseekv4}        & ---          & ---          & 4.77 & 4.57 & 4.43 \\
Gemini~\cite{gemini31pro}  & 4.10 & 4.61 & 4.86 & 4.94 & 4.16 \\
\bottomrule
\end{tabular}%
}
\end{table}

\begin{table}[htb]
\centering
\small
\caption{Comparison with prior corpora: \textbf{Dur. (hr)} total audio duration in hours; \textbf{Type} (natural / synthesized); \ding{172} persona + scenario; \ding{173} full-duplex status; \ding{174} sound event. SDF abbreviates SpeechDialogueFactory. Legend: \ding{55} absent, $\checkmark$ present.}
\label{tab:comparison}
\begin{tabular}{l ccccc}
\toprule
\textbf{Dataset} & \textbf{Dur. (hr)} & \textbf{Type} & \ding{172} & \ding{173} & \ding{174} \\
\midrule
Fisher~\cite{fisher}                   & 2{,}000  & natural & \ding{55} & \ding{55} & \ding{55} \\
CANDOR~\cite{candor}                  & ~850     & natural & \ding{55} & \ding{55} & \ding{55} \\
Open-Yap-1K~\cite{openyap1k}            & 1{,}000  & natural & \ding{55} & \ding{55} & \ding{55} \\
DuplexConv~\cite{duplexconv}            & ~2{,}000  & natural & \ding{55} & $\checkmark$ & $\checkmark$ \\
SDF~\cite{sdf}                         & ~146     & synth.  & $\checkmark$ & \ding{55} & \ding{55} \\
\midrule
\textbf{Ours}               & 800      & synth.  & $\checkmark$ & $\checkmark$ & $\checkmark$ \\
\bottomrule
\end{tabular}
\end{table}

\subsection{Comparison with Other Datasets}
\label{ssec:comparison}

Table~\ref{tab:comparison} compares our released DuplexDrama with five prior corpora. \emph{Fisher}~\cite{fisher} and \emph{CANDOR}~\cite{candor} supply natural conversational speech as acoustic baselines but ship without any of our four annotation dimensions. \emph{Open-Yap-1K}~\cite{openyap1k} and \emph{DuplexConv}~\cite{duplexconv} ship large-scale natural conversational recordings with speaker overlap, yet both lack scripted scenarios, persona attributes, and sound-event annotations. \emph{SpeechDialogueFactory}~\cite{sdf} is a synthesized corpus that provides persona and scenario annotations but does not cover full-duplex scenarios. Our proposed DuplexDrama is the only corpus that jointly covers all four annotation dimensions, with the script-aware sound event tags being unique to our work.

\section{Conclusion}
\label{sec:conclusion}

We present DuplexDrama, the first TTS-synthesized spoken dialogue
dataset jointly covering persona and scenario annotations,
full-duplex behaviors, expressive speech, and sound events.
The corpus is built through a four-stage pipeline and evaluated along
four objective audio metrics, with dual-LLM cross-checking of script
rationality. We will release
approximately 800 hours of data to facilitate research on full-duplex
spoken dialogue modeling.
Future work will pursue more realistic duplex label distributions,
paralinguistic phenomena in TTS output, and more flexible sound-event
label matching.

\section*{Acknowledgement}
\textit{AI Disclosure.} LLMs were used solely to polish English during
manuscript preparation; all technical content is authored and verified
by the human authors, who take full responsibility for the final text.
\textit{Compliance with Ethical Standards.} DuplexDrama is constructed
from TTS synthesis and publicly available audio; no human subject data
was used; ethical approval was not required.
\textit{Conflicts of Interest.} None.


\bibliographystyle{IEEEbib}
\bibliography{strings,refs}

\end{document}